\documentclass{article}

\usepackage{arxiv}

\usepackage[utf8]{inputenc} % allow utf-8 input
\usepackage[T1]{fontenc}    % use 8-bit T1 fonts
\usepackage[numbers]{natbib}% citations (must load before hyperref)
\usepackage{hyperref}       % hyperlinks
\usepackage{url}            % simple URL typesetting
\usepackage{booktabs}       % professional-quality tables
\usepackage{amsfonts}       % blackboard math symbols
\usepackage{amsmath}        % align environment
\usepackage{nicefrac}       % compact symbols for 1/2, etc.
\usepackage{microtype}      % microtypography
\usepackage{graphicx}       % figures

\title{Investigating the Ability of Large Language Models to Analyze Recipes for Diabetes}

\author{
  Revathy Venkataramanan,
  Aditya Luthra,
  Venkatesan Nadimuthu,
  Amit Sheth \\
  AI Institute, University of South Carolina \\
  1112 Greene Street, Columbia, SC 29201 USA \\
  \texttt{revathycv24@gmail.com} \\
}
\date{}

\renewcommand{\undertitle}{Preprint}
\renewcommand{\headeright}{Preprint}
\renewcommand{\shorttitle}{LLMs for Analyzing Recipes for Diabetes}

\begin{document}
\maketitle

\begin{abstract}
Several studies have evaluated the ability of Large Language Models (LLMs) for meal planning, yielding positive outcomes. These models can process natural language inputs and leverage learned knowledge from their pretraining to generate meal plans. In this work, we investigate the ability of LLMs to analyze the suitability of given recipes for diabetes. The primary challenge for LLMs is to retrieve relevant dietary guidelines for diabetes, decompose recipes into ingredients and cooking methods, and apply these guidelines to determine the recipe's suitability. To study these challenges, we employ three kinds of prompts namely, (i) Direct Query Prompt (ii) Context-Guided Prompt, and (iii) Exemplary Context Prompt that incorporate different levels of diabetes dietary guidelines from medical sources. We introduce a benchmark dataset curated for this investigation consisting of 7607 recipes that include 3807 recipes suitable for diabetes and 3800 recipes not suitable for diabetes. Our results demonstrate that most LLMs are cautious in predicting recipes as suitable to prevent detrimental outcomes. Further, the models that can reason using the dietary guidelines performed better in predicting the suitability of recipes for diabetes. Overall, Mistral-7B and Llama 70B showed superior performance to their counterparts.
\end{abstract}


%========================= Introduction =========================
\section{Introduction}\label{sec:introduction}

The World Health Organization estimates that 830 million people worldwide have diabetes \cite{WHO2024}, a condition manageable through dietary adjustments. Comprehensive dietary guidelines have been published by medical sources like MayoClinic and the Center for Disease Control (CDC). However, they are extensive, challenging to remember, and difficult to apply in daily meal planning. While the guidelines include examples, they are not exhaustive. For instance, vegetables and fruits are considered nutritious sources of healthy carbohydrates and rich fiber, suitable for diabetes management. However, certain fruits high in glucose should be avoided. This underscores the complexity of dietary decisions for diabetes. While AI-powered meal-tracking tools \cite{min2019survey} have gained popularity over the years, they often lack targeted analysis for diabetes, likely due to liability concerns.

General-purpose LLMs such as ChatGPT, Bard, and Claude have become invaluable tools for the general public. They are frequently employed for everyday activities like meal planning \cite{Cesaric2024}, vacation itineraries \cite{shin2023chatgpt}, drafting emails, and providing educational assistance. There have been several studies that showcase the positive aspect of LLMs in meal planning \cite{petruzzelli2024recommending}. At the same time, they are suggested to be augmented with the clinician's advice \cite{ponzo2024chatgpt}. On the other hand, studies have also shown the limitations of using LLMs for nutrition management, specifically for diabetes \cite{medicalxpress2024ai, naja2024artificial}.

In this study, we aim to investigate the ability of LLMs to assess the suitability of a given recipe for diabetes. Unlike meal planning, where LLMs can take a cautious approach by recommending meals derived using a subset of diabetes guidelines that they can retrieve or remember, assessing suitability demands a thorough understanding. This process requires comprehensive knowledge of the full scope of dietary guidelines. The core challenges for the LLMs include \textit{\textbf{(i) Medical Knowledge Retrieval:}} Searching their vast embedding space for relevant, targeted and complete dietary guidelines for diabetes.
\textit{\textbf{(ii) Conceptual Understanding:}} Interpreting diabetes dietary concepts (what constitutes healthy carbohydrate or saturated fat).
\textit{\textbf{(iii) Deductive Analysis:}} Applying general dietary rules to specific cases (e.g., ``Carrot is a vegetable, vegetables are healthy carbohydrates'') to analyze suitability. For this investigation, we utilize three types of prompts incorporating different levels of diabetic dietary guidelines: \textit{\textbf{(i) Direct Query Prompt (ii) Context-Guided Prompt, and (iii) Exemplary Context Prompt}}. We further prompt the LLMs to reason using the keywords part of dietary guidelines such as healthy carbohydrates, trans fat, and so on. Four classes of LLMs namely, Mistral, Gemma2, Llama, and ChatGPT with varying sizes were chosen for this analysis. To conduct this investigation, we introduce a recipe dataset consisting of 7607 recipes with 3807 recipes suitable for diabetes gathered from medical sources and 3800 recipes not suitable for diabetes curated using keywords on Recipe1M dataset \cite{salvador2017learning}.\footnote{The code and dataset are available at \url{https://github.com/revathyramanan/LLMs-Diabetes-Recipes}.}

%========================= Related Work =========================
\section{Related Work}\label{sec:related_work}

LLMs are being explored for generating personalized diets aligned with nutritional guidelines, focusing on managing conditions like diabetes.

\subsection{LLMs in Diet Management}
Recent advances in LLMs have transformed food recommendation systems \cite{moskowitz2024foods}. \cite{petruzzelli2024recommending} developed a framework that identifies healthier and more sustainable meals by exploiting retrieval techniques and large language models. \cite{RW-Zhou_2024} introduced FoodSky, a food-specific LLM with enhanced reasoning for dietary tasks. \cite{RW-Yang_2024} developed ChatDiet, a chatbot combining LLMs with causal reasoning for personalized recommendations. \cite{RW-Chen_2023} used multi-task learning with knowledge graphs to balance user preferences and health needs. \cite{RW-Szymanski_2024} developed a template-based GPT-4 nutrition assistant, validated with dietitian input for accuracy and relevance. \cite{RW-Rodriguez_2023} introduced ``Dining on Details,'' a framework using LLMs and multi-modal approaches for fine-grained food recognition in chronic disease dietary monitoring. LLMs are increasingly utilized for dietary personalization. \cite{RW-Papastratis_2024} combined ChatGPT with a generative model to provide robust meal plans aligned with established nutritional guidelines.

\subsection{LLMs for Disease Diet Management}
LLMs have been employed and evaluated for generating meal plans. \cite{RW-Abbasian_2024} developed a knowledge-infused conversational health agent that integrates dietary guidelines for diabetes management, outperforming general LLMs in reliability. \cite{RW-Bungay_2024} focused on diabetic nutrition with DIAFOODS, a tool that personalizes dietary recommendations to optimize glycemic control. Additionally, \cite{RW-Min_2023} demonstrated the role of structured nutritional strategies in managing Type 2 Diabetes Mellitus, emphasizing individualization. \cite{RW-Sun_2023} introduced an AI dietitian leveraging LLMs and image recognition for managing diabetes, which integrates ingredient identification and meal assessments. \cite{RW-Wang_2024} introduced the RISE framework, which augments LLMs with retrieval systems to provide more accurate and comprehensible responses to diabetes-related queries. LLMs have also been evaluated for meal planning in other conditions such as gastrointestinal \cite{lamb2024s2185}, Mediterranean diet \cite{chen2024evaluation} and non-communicable diseases \cite{papastratis2024can}.

%========================= Approach =========================
\section{Approach}\label{sec:approach}

\subsection{Dataset Creation}
\paragraph{(i) Diabetes Specific Recipes:}
Several reliable medical websites provide curated lists of recipes designed specifically for individuals managing diabetes. These recipes serve as practical guides to help diabetes patients control their condition through healthy dietary choices. In our work, we utilized MayoClinic \cite{mayoclinic2024diabetes}, Diabetes UK \cite{diabetesuk2024recipes} and Diabetes Hub \cite{diabetesfoodhub2024recipes} to compile a collection of 3,807 recipes. This approach ensures that the dataset is grounded in medical expertise, supporting its relevance and usefulness for dietary management research.

\paragraph{(ii) Non-Diabetes Specific Recipes:}
MayoClinic \cite{mayoclinic2024diabetesdiet} provides comprehensive guidelines for diabetes patients, detailing both recommended and discouraged food items. For example, foods high in refined sugars and saturated fats are advised against, while high-fiber vegetables and whole grains are encouraged. We curated a set of keywords listed by MayoClinic to identify recipes that may not be suitable for diabetes. The keywords include `ribs', `pork bacon', `sausage', `baked', `shortening', `pancakes', `cookies', `tart', `pudding', `pickles', `soda', `syrup', `churro', and a list of canned fruits and vegetables. Leveraging the Recipe1M dataset \cite{salvador2017learning}, we conducted keyword searches within recipe titles and ingredients. To ensure class balance, we randomly selected 3800 recipes from the result to construct the dataset.

\renewcommand{\arraystretch}{1.5}
\begin{table}[t]
\centering
\setlength{\tabcolsep}{4pt}
\caption{Performance results of LLMs in analyzing the suitability of recipes for diabetes}
\label{table:metrics_all}
\begin{tabular}{cccccccccc}
\hline
 & \multicolumn{3}{c}{Prompt-1} & \multicolumn{3}{c}{Prompt-2} & \multicolumn{3}{c}{Prompt-3} \\
 \hline
Models & Accuracy & Precision & Recall & Accuracy & Precision & Recall  & Accuracy & Precision & Recall \\
\hline
Mistral 7B & \textbf{0.84} & 0.84 & \textbf{0.84} & 0.77 & 0.88 & 0.63 & 0.79 & 0.54 & 0.53 \\
Mistral 12B & 0.60 & 0.92 & 0.21 & 0.62 & 0.67 & 0.45 & 0.63 & 0.74 & 0.41 \\
Gemma2 2B & 0.51 & 0.47 & 0.34 & 0.78 & 0.86 & 0.67 & 0.75 & 0.89 & 0.58 \\
Gemma2 9B & 0.70 & 0.53 & 0.47 & 0.76 & \textbf{0.96} & 0.54 & 0.81 & 0.95 & 0.65 \\
Gemma2 27B & 0.83 & 0.94 & 0.71 & 0.78 & \textbf{0.96} & 0.59 & 0.79 & \textbf{0.96} & 0.61 \\
Llama3.1 8B & 0.61 & 0.90 & 0.25 & 0.72 & 0.90 & 0.50 & 0.81 & 0.90 & 0.70 \\
Llama3.1 70B & 0.69 & \textbf{0.96} & 0.40 & \textbf{0.83} & 0.92 & \textbf{0.72} & \textbf{0.85} & 0.91 & \textbf{0.79} \\
Llama3.2 2B & 0.56 & 0.64 & 0.29 & 0.51 & 0.95 & 0.03 & 0.54 & 0.90 & 0.09 \\
Chatgpt 3.5 & 0.74 & 0.50 & 0.49 & 0.54 & 0.90 & 0.09 & 0.74 & 0.50 & 0.49 \\
\hline
\end{tabular}
\end{table}

\subsection{Model Details}
Four classes of LLMs namely, Mistral, Llama, Gemma2, and ChatGPT with small (less than 3 billion), medium (between 3 and 9 billion), and large (greater than 9 billion) sizes were chosen for this evaluation. The models are (i) Mistral 7B and Mistral 12B \cite{jiang2023mistral}, a model leveraging transformers architecture with sliding window attention and pre-fill chunking, (ii) Gemma2 2B, Gemma2 9B and Gemma2 27B, light-weight models with local-global attentions and group-query attention \cite{team2024gemma} (iii) Llama 3.1 8B, Llama 3.1 70B, Llama 3.2 3B, models with SwiGLU activation function and Rotary Embeddings \cite{touvron2023llama} (iv) ChatGPT-3.5.

\subsection{Prompting Techniques}
To evaluate the core challenges presented, we employ three types of prompt to investigate the ability of LLMs to analyze the suitability of recipe for diabetes.

\subsubsection{Direct Query Prompt (Prompt-1):}
In this approach, a straightforward, single-line query is presented alongside the recipe. This aims to evaluate whether LLMs can recognize that the query pertains to a diabetes-specific diet and retrieve relevant medical guidelines from their vast embedding space learned during their training to analyze meals by reasoning over recipe contents. This refers to \textit{medical knowledge retrieval.}

\begin{quote}
\textit{
Give one word Answer YES/NO.
Is the Given Recipe safe for diabetes?
The Title of the Recipe is:
\{recipe title\}.
The Ingredients of \{recipe title\} are:
\{comma separated list of ingredients\}.
The Instructions of \{recipe title\} are:
\{list of instructions\}.}
\end{quote}

\subsubsection{Context Guided Prompt (Prompt-2):}
This prompt incorporates explicit dietary guidelines. Though models can utilize this knowledge instead of retrieving from its search space, this requires a \textit{conceptual understanding} what are healthy carbohydrates, saturated fat and other dietary concepts. This is essential to apply this knowledge to individual recipe ingredients to analyze their suitability. The guidelines were curated from MayoClinic \cite{mayoclinic2024diabetesdiet} and NIDDK \cite{niddk2024diabetes}, which outline the recommended foods and those to avoid. This also evaluates if the LLMs prioritize the external knowledge provided over its internal knowledge.

\begin{quote}
    \textit{
    You will be given a recipe with title, list of ingredients and instruction to make the recipe. You will also be given dietary guidelines for diabetes that tells items to recommend and avoid. Now, based on this information, tell me in one word if the recipe given is suitable for diabetes or not using YES or NO options. Then provide 2-3 line reasoning along with YES or NO as why a recipe is suitable or not suitable using those keywords in the given dietary guidelines.} \\
    \textit{
    Recipe Title: \{add recipe title\} \\
    Recipe Ingredients: \{list of ingredients separated by comma\} \\
    Recipe Instructions: \{a paragraph of instructions\}}

    \textit{Diabetes Dietary Guidelines:
    For diabetes, the recommended items are (i) Healthy Carbohydrates (ii) Fibre rich foods (iii) Heart Healthy fish (iv) Good Fats (v) Animal Protein (vi) Protein (vii) Non fat, Low fat.
    The items that should be avoided are (i) Saturated fat (ii) High dairy fat (iii) Saturated fat, animal protein (iv) Trans fat (v) Cholesterol (vi) Added sugar (vii) Refined/processed carbohydrates (viii) Ultra processed foods (ix) deep fried foods}
\end{quote}

\subsubsection{Exemplary Context Prompt (Prompt-3):}
This prompt is an enhanced version of the Context Guided Prompt, that includes specific examples for dietary concepts. This approach aims to achieve a balance between providing direct and indirect examples. For instance, vegetables are included as examples of healthy carbohydrates within the prompt without explicitly labeling whether an ingredient is a vegetable. In contrast, categories such as saturated fats include direct examples like butter, beef, and sausages. This prompt assesses how effectively LLMs can perform \textit{deductive analysis}. For example, deducing that carrot is a vegetable and vegetable are healthy carbohydrates. Further, this evaluation seeks to determine whether the explicit inclusion of such comprehensive information enhances the LLMs' performance.

\begin{quote}
    \textit{
    You will be given a recipe with title, list of ingredients and instruction to make the recipe. You will also be given dietary guidelines for diabetes that tells items to recommend and avoid. Now, based on this information, tell me in one word if the recipe given is suitable for diabetes or not using YES or NO options. Then provide 2-3 line reasoning along with YES or NO as why a recipe is suitable or not suitable using those keywords in the given dietary guidelines.} \\
    \textit{
    Recipe Title: \{add recipe title\} \\
    Recipe Ingredients: \{list of ingredients separated by comma\} \\
    Recipe Instructions: \{a paragraph of instructions\}}

    \textit{Diabetes Disease Context:
    For diabetes, the recommended items are
    (i) Healthy Carbohydrates such as fruits, vegetables, whole grains, legumes and non-fat dairy
    (ii) Fibre-rich foods such as fruits, vegetables, nuts, legumes, and whole grains
    (iii) Heart Healthy fish such as salmon, tuna, mackerel and sardines
    (iv) Good Fats such as avocado, nuts, olive oil, canola oil and peanut oil
    (v) Animal Protein such as lean meat, chicken without skin, turkey without skin, fish and eggs
    (vi) Protein such as nuts, peanuts, dried beans, chickpeas, split peas and tofu
    (vii) Non-fat, Low fat such as milk, lactose-free milk, yogurt, and cheese}
    \textit{
    The items that should be avoided are
    (i) Saturated fats such as fatty meat, pork, eggs, palm oil, coconut oil and red meat
    (ii) High dairy fat such as butter
    (iii) Saturated fat, animal protein such as beef, hot dogs, sausage and bacon
    (iv) Trans fat such as processed snacks, baked goods, shortening, stick margarine, refrigerated doughs, pastries, cookies, crackers, pies, and stick butter replacements
    (v) Cholesterol such as high-fat dairy, high-fat animal protein, egg yolks, liver, organ meat
    (vi) Added sugar such as soda, sports drinks, energy drinks, candy, canned fruits, jams, jellies and ice cream
    (vii) Refined/processed carbohydrates such as white bread, pasta, juice, sweets, candy, cake
    (viii) Ultra processed foods such as hot dogs, fried fish sticks and fast food burgers
    (ix) deep fried foods such as potato fries and chicken nuggets}
\end{quote}

\subsection{Experimentation}
Each pretrained model was downloaded and run using 8xV100-40GB machines. Each model was prompted with one recipe at a time individually for each prompt. The results were stored along with explanations and reasoning provided by LLMs. For prompt-2 and prompt-3, the models were prompted to provide reasons for their decision using the concepts given in the dietary guidelines. Chatgpt-3.5 was accessed using its API.

%========================= Results and Discussion =========================
\begin{figure}[t]
\centering
\includegraphics[width=\textwidth]{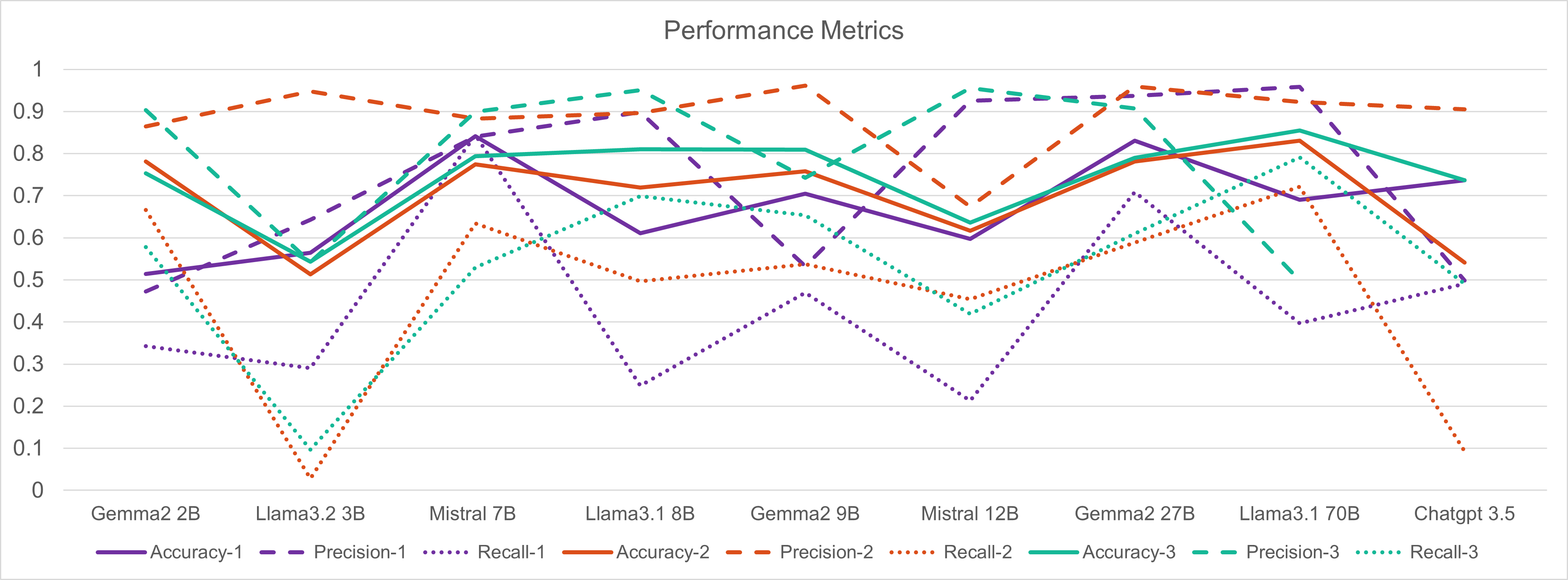}
\caption{Performance metrics of all the LLMs across all three prompts visualized to study correlations. The number -1, -2 and -3 denote results of prompt-1, prompt-2 and prompt-3}
\label{fig:chart_metrics}
\end{figure}

\section{Result and Discussion}\label{sec:result_discussion}

\paragraph{Model Performance:}
Table \ref{table:metrics_all} provides the performance metrics for all models across the three prompt types: Direct Query Prompt (Prompt-1), Context Guided Prompt (Prompt-2), and Exemplary Context Prompt (Prompt-3). Mistral 7B and Llama3.1 70B exhibit consistent performance with minimal variation between precision and recall across all prompts compared to other models. Mistral-7B performed well in Direct Query Prompt, and Llama3.1-70B in Context Guided Prompt and Exemplary Prompt. Llama3.1 8B, a medium-sized model also seems to have benefited from Exemplary Context Prompt. While most Gemma2 class models have high precision, the recall is low. As shown in Figure \ref{fig:chart_metrics}, recall (dotted line) tends to be lower than precision (dashed line) across all prompts for most models. This indicates that the LLMs may be biased toward predicting recipes as ``not suitable for diabetes'' to avoid false positive errors. Consequently, the models appear overly cautious in classifying recipes as ``suitable.'' In this context, the cost of false positives, incorrectly identifying a harmful recipe as safe for diabetes, could have adverse outcomes and the models seem to be aware of it.

\paragraph{Precision-Recall Stability Metric:}
Intuitively, significant variation between precision and recall suggests that the model favors one class over the other. To quantify the variation between high precision and low recall, we proposed a stability metric in Equation \ref{eq:stability}. A higher stability score indicates lower variation, signifying a more balanced and stable model. A score of 1 represents a highly stable model, and a score of 0 indicates instability.

\begin{align}
\label{eq:stability}
    Stability = 1 - |Precision-Recall|
\end{align}

\paragraph{Impact of Reasoning on Model Performance and Stability:}
In Context Guided Prompt (prompt-2) and Exemplary Context Prompt (prompt-3), models were instructed to justify their decisions using keywords derived from dietary guidelines, such as Healthy Carbohydrates, Trans Fat, Cholesterol, and others. Since dietary guidelines were not incorporated into the Direct Query Prompt (prompt-1), models were not instructed to provide reasoning. Gemma2 27B, a lightweight faster model, was used to extract keywords from the reasoning paragraph provided by LLMs for each recipes. The results showed that models classified a recipe as suitable for diabetes justifying by leveraging keywords from both the ``recommend'' and ``avoid'' sections of the guidelines. For example, a model might justify its prediction by stating that a recipe lacks saturated and trans fats, aligning with the ``recommend'' criteria. Consequently, when the model predicted a positive outcome (suitable), we utilized keywords from the ``recommend'' section for validation. Conversely, for negative predictions (not suitable), we extracted relevant keywords from the ``avoid'' section to evaluate the model's rationale.

\begin{figure}[t]
\centering
\includegraphics[width=\linewidth]{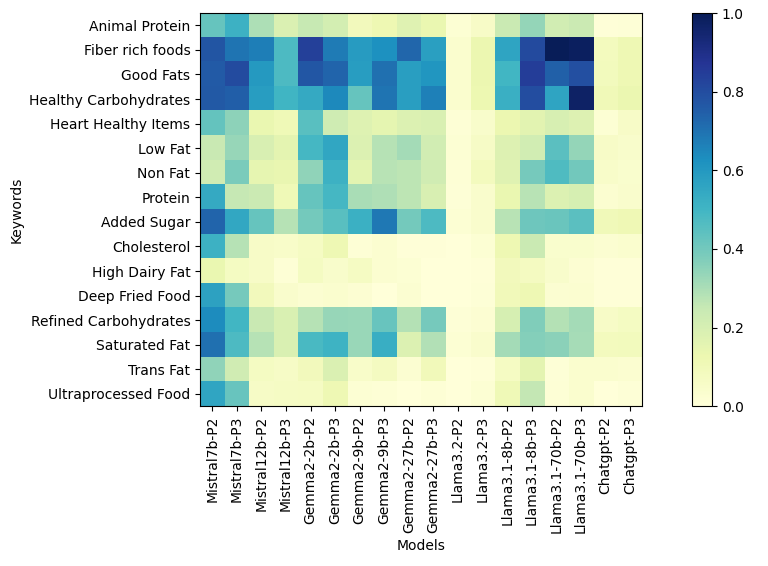}
\caption{Keywords from diabetes dietary guidelines used by the models to reason over their decision making. The values are normalized using min-max scaling. P2 denote prompt-2 and P3 denote prompt-3}
\label{fig:heatmap}
\end{figure}

\begin{figure}[t]
\centering
\includegraphics[width=\linewidth]{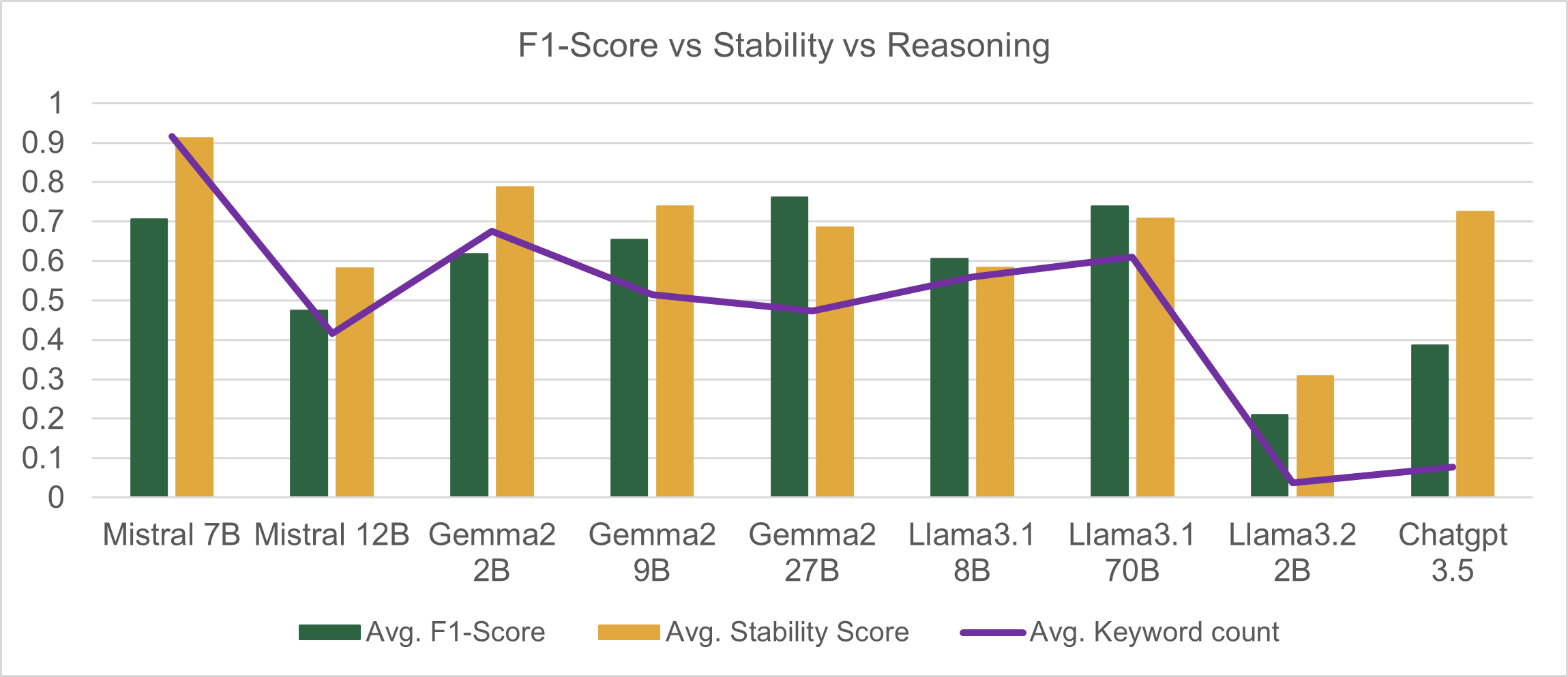}
\caption{Correlation graph illustrating the relationship between models' reasoning ability and their performance and stability.}
\label{fig:keyword_performance}
\end{figure}

The heatmap presented in Figure \ref{fig:heatmap} shows Mistral 7B produced the highest amount of diverse dietary guideline keywords in its reasoning. This supports the results presented in Table \ref{table:metrics_all} where Mistral 7B showed steady results across all prompts with less variation between precision and recall. Llama3.1 70B returned keywords clustered around good fats, healthy carbohydrates, and fibre-rich foods. To correlate better, we computed the average F1-score, average stability score (Equation \ref{eq:stability}), and the average count of keywords across all prompts for a given model. Figure \ref{fig:keyword_performance} demonstrates that the more keywords produced by the model in its reasoning, the better the F1-score and/or better the stability (less variation between precision and recall). It is to be noted that Gemma2-2B is the model with the second-highest stability score that correlates with the second-highest amount of keywords provided in the reasoning. Llama3.1-70B has the third-highest average of keyword count. ChatGPT and Llama3.2-2B had the least amount of keywords in their reasoning which correlates with lower F1-score. If a model confidently reasons with sufficient concepts from dietary guidelines, the model's stability and performance is better.

\paragraph{Deductive Analysis:}
It is evident from Figure \ref{fig:keyword_performance} that if a model learned to apply the deductions and provide reasoning using given dietary guidelines, the better the performance. For example, if a model can deduce $carrot \rightarrow vegetable$ and $vegetable \rightarrow healthy\ carbohydrate$ and $healthy\ carbohydrates \rightarrow suitable\ for\ diabetes$, the model analyzes the recipes better. This shows that the ability to reason over its decisions improves the model's stability and performance. However, we cannot say for certain that the models did such detailed reasoning over individual ingredients, which we aim to evaluate in our future work. The individual F1 scores, stability score, and keyword counts can be found here - \textbf{\url{https://tinyurl.com/ycxf48bz}}.\footnote{Code and dataset: \url{https://github.com/revathyramanan/LLMs-Diabetes-Recipes}.}

\paragraph{Medical Knowledge Retrieval:}
Prompt-1 requires models to retrieve dietary guidelines from their internal embedding space to analyze the recipe's suitability. To assess if the retrieved knowledge is comprehensive, we re-ran experiments for prompt-1 instructing the models to reason over their decision-making. Mistral 7B, Gemma 27B, and Llama3.1 70B, one model from each class were chosen to comparatively study the reasoning trends for all three prompts. Results showed prompt-1 produced fewer average keywords (765) than prompt-2 (836) and prompt-3 (789). Models retrieved internal medical knowledge, but external prompts provided easier access resulting in better performance supported by reasoning. However, for ChatGPT, prompt-2's external knowledge conflicted with internal knowledge, yielding high precision but low recall.

\paragraph{Conceptual Understanding:}
Figure \ref{fig:heatmap} highlights that models confidently predict concepts like fiber-rich foods, good fats, healthy carbohydrates, refined carbohydrates, and added sugar demonstrating a solid understanding of these areas. Because of this, LLMs may have exhibited positive outcomes in meal planning by suggesting meals that confirm to these concepts. However, they perform less effectively with categories such as non-fat, low-fat, protein, cholesterol, high dairy fat, trans fats, deep-fried foods, and ultra-processed foods. Deep-fried food and trans-fat food may require analyzing cooking methods. This underscores the complexity of recipe evaluation, as it demands assessing both ingredients and cooking methods against dietary guidelines individually which is challenging for general-purpose LLMs \cite{dziri2024faith}.

\paragraph{Deep Dive into Mistral-7B:}
Figure \ref{fig:chart_metrics} shows that larger models do not necessarily guarantee better performance. Mistral 7B, which generally prioritizes and aligns with given external knowledge \cite{jiang2023mistral}, demonstrated superior reasoning using dietary guideline keywords. It averaged 1200 keywords for prompt-2 and 1008 for prompt-3, the highest among all models. The decrease in keywords from prompt-2 to prompt-3 corresponds to a drop in performance as well (Table \ref{table:metrics_all}), reinforcing the correlation between reasoning using keywords and model performance. The performance decline in prompt-3 may be due to prompt length or insufficient examples of dietary concepts like trans fat, as Mistral 7B heavily relies on external knowledge despite its internal understanding.

\paragraph{Curious Case of ChatGPT:}
The precision and recall did not vary for prompt-1 and prompt-3 for ChatGPT. In other words, the model was stable even though the accuracy was low. However, for prompt-2, which included only diabetes dietary concepts and not examples, the internal knowledge of ChatGPT seems to have conflicted with the external knowledge given. A well-recognized issue \cite{zhang2024evaluating} that resulted in a low recall. However, the examples given in prompt-3 improved the recall. Going against the well-known trend that bigger models perform better, ChatGPT showed inconsistencies due to conflicting knowledge. To add, a bigger embedding search space makes it difficult for the models to retrieve targeted medical advice \cite{huang2024tool}.

\subsection{Technical Challenges}
Revisiting the challenges presented in the introduction, \textit{\textbf{(i) Medical Knowledge Retrieval:}} The results of prompt-1 show that the models trained for general purpose find it challenging to search their vast embedding space for targeted relevant dietary guidelines compared to explicitly providing knowledge via prompts. Complete and comprehensive retrieval of medical knowledge is critical in this domain. \textbf{\textit{(ii) Conceptual Understanding:}} The models seem to have a good understanding of straightforward concepts such as healthy carbohydrates, fiber-rich foods, added sugar, saturated fats, and refined carbohydrates. However, a thorough understanding of all the concepts is essential to make informed decisions. Deep-fried food requires knowledge about cooking methods as well. \textbf{\textit{(iii) Deductive Analysis:}} To a certain extent, we can say that the models were able to deduce and reason using dietary concepts. However, it is not certain if the models individually applied reasoning over ingredients and cooking methods to derive conclusions. In our future work, we aim to evaluate the ability of LLMs to decompose a recipe into ingredients and cooking methods, reason over each ingredient and cooking method in a recipe to evaluate if LLMs can perform compositional reasoning \cite{dziri2024faith}.

\subsection{Clinical Challenges}
\textbf{\textit{(i) Alignment to Medical Guidelines:}} While analyzing and reasoning, consistent alignment with medical guidelines is critical. This is a challenge as LLMs are known to hallucinate \cite{rawte2023survey}.
\textbf{\textit{(ii) Knowledge Limitation:}} Diabetes guidelines require knowledge of nutrition, glycemic index, and other factors, making it challenging to integrate this information from vast search spaces.
\textbf{\textit{(iii) Expert Validation:}} The model needs to be validated by domain experts which involves tedious human labor.

\subsection{Ethical Challenges}
\textbf{\textit{(i) Safety:}} Any incorrect or misleading information generated by LLMs poses significant risks, particularly in a high-stakes domain such as diet and chronic conditions. It could lead to adverse health outcomes if this advice is blindly followed.
\textit{\textbf{(ii) Biased Representation:}} There is an unequal representation of dietary guidelines, as only a few categories are used as reasoning despite having several categories to recommend or avoid.

%========================= Conclusion =========================
\section{Conclusion}\label{sec:conclusion}
In this work, we investigated the ability of LLMs to analyze the suitability of a given recipe for diabetes. Unlike meal planning, where LLMs can take a cautious approach by recommending meals derived using guidelines that they can retrieve or remember, assessing suitability demands a full scope of guidelines. Mistral 7B, Llama3.1-70B showed superior performance overall to its counterparts. The Gemma2 class of models had high precision and low recall. Most models were conservative in predicting the recipes suitable for diabetes due to the high-stakes nature of this domain. Of the core challenges discussed, most models were able to retrieve the medical knowledge from their vast embedding space. However, better performance was shown when external knowledge was given in the prompt. Conversely, this external knowledge also conflicted with the internal knowledge of the model in the case of ChatGPT. The models had a good understanding of concepts such as healthy carbohydrates, good fat, added sugar, and refined carbohydrates. On the other hand, models need to understand and utilize categories such as protein, non-fat, low-fat, deep-fried food, and ultra-processed food in their reasoning. In addition, the models that were able to deductively analyze and reason using dietary concepts showed better performance. In our future work, we aim to evaluate if the LLMs can individually reason over ingredients and cooking methods to drive a conclusion on the suitability of recipes for diabetes.

%========================= References =========================
\bibliographystyle{unsrtnat}
\bibliography{references}

\end{document}